\documentclass[11pt,a4paper]{article}
\usepackage[hyperref]{emnlp2018}
\usepackage{times}
\usepackage{latexsym}
\usepackage{url}

\usepackage{graphicx}
\usepackage{amsmath}
\usepackage{multirow}
\usepackage{xspace}
\usepackage{bm}
\usepackage{amssymb}
\usepackage{booktabs}
\usepackage{pgfplots}

\newcommand{\secref}[1]{Section~\ref{#1}\xspace}
\newcommand{\tabref}[2][]{Table#1~\ref{#2}\xspace}
\newcommand{\figref}[2][]{Figure#1~\ref{#2}\xspace}

\newcommand{\myparagraph}[1]{\noindent{\textbf{#1}}~}

\newcommand{\mat}[2][]{\boldsymbol{#2}_{#1}}

\renewcommand{\vec}[2][]{\boldsymbol{#2}_{#1}}
\newcommand{\vecfwd}[2][]{\overrightarrow{\bm{#2}}_{#1}}
\newcommand{\vecbwd}[2][]{\overleftarrow{\bm{#2}}_{#1}}

\newcommand{\lstmfwd}{\overrightarrow{\textrm{LSTM}}}
\newcommand{\lstmbwd}{\overleftarrow{\textrm{LSTM}}}

\newcommand{\xentropy}{\ensuremath{\operatorname{XEntropy}}\xspace}

\newcommand*{\affmark}[1][*]{\textsuperscript{#1}}

\aclfinalcopy % Uncomment this line for the final submission
\def\aclpaperid{1518} %  Enter the acl Paper ID here

\title{Evaluating the Utility of Hand-crafted Features in Sequence Labelling\thanks{{} {} \url{https://github.com/minghao-wu/CRF-AE}}}
\author{Minghao Wu\affmark[$\spadesuit\heartsuit$]\thanks{{} {} Work carried out at The University of Melbourne} \qquad Fei Liu\affmark[$\spadesuit$] \qquad Trevor Cohn\affmark[$\spadesuit$]\\
         \affmark[$\spadesuit$]The University of Melbourne, Victoria, Australia\\
         \affmark[$\heartsuit$]JD AI Research, Beijing, China \\
         {\tt {wuminghao@jd.com}} \\ % TODO: please verify this
         {\tt {fliu3@student.unimelb.edu.au}}\\
         {\tt {t.cohn@unimelb.edu.au}}}

\date{}

\begin{document}
\maketitle
\begin{abstract}
Conventional wisdom is that hand-crafted features are redundant for deep learning models, as they already learn adequate representations of text automatically from corpora. 
In this work, we test this claim by proposing a new method for exploiting handcrafted features as part of a novel hybrid learning approach, incorporating a feature auto-encoder loss component. 
We evaluate  on the task of named entity recognition (NER), where we show that including manual features for part-of-speech, word shapes and gazetteers can improve the performance of a neural CRF model.
We obtain a $F_1$ of 91.89 for the CoNLL-2003 English shared task, which significantly outperforms a collection of highly competitive baseline models. 
We also present an ablation study showing the importance of auto-encoding, over using features as either inputs or outputs alone, and moreover, show including the autoencoder components reduces training requirements to 60\%, while retaining the same predictive accuracy.

\end{abstract}

\section{Introduction}
Deep neural networks have been proven to be a powerful framework for natural language processing, and have demonstrated strong performance on a number of challenging tasks, ranging from machine translation \cite{cho2014learning,cho2014properties}, to text categorisation \cite{zhang2015character, joulin2017bag, Liu+:2018b}. 
Not only do such deep models outperform traditional machine learning methods, they also come with the benefit of not requiring difficult feature engineering. 
For instance, both \newcite{lample2016neural} and \newcite{ma2016end} propose end-to-end models for sequence labelling task and achieve state-of-the-art results. 

Orthogonal to the advances in deep learning is the effort spent on feature engineering. 
A representative example is the task of named entity recognition (NER), one that requires both lexical and syntactic knowledge, where, until recently, most models heavily rely on statistical sequential labelling models taking in manually engineered features \cite{florian2003named,chieu2002named,ando2005framework}. 
Typical features include POS and chunk tags, prefixes and suffixes, and external gazetteers, all of which represent years of accumulated knowledge in the field of computational linguistics.

The work of \newcite{collobert2011natural} started the trend of feature engineering-free modelling by learning internal representations of compositional components of text (e.g., word embeddings). 
Subsequent work has shown impressive progress through capturing syntactic and semantic knowledge with dense real-valued vectors trained on large unannotated corpora \cite{Mikolov+:2013a,Mikolov+:2013b,pennington2014glove}.
Enabled by the powerful representational capacity of such embeddings and neural networks, feature engineering has largely been replaced with taking off-the-shelf pre-trained word embeddings as input, thereby making models fully end-to-end and the research focus has shifted to neural network architecture engineering.

More recently, there has been increasing recognition of the utility of linguistic features \cite{Li+:2017,Chen+:2017,Wu+:2017,Liu+:2018} where such features are integrated to improve model performance. 
Inspired by this, taking NER as a case study, we investigate the utility of hand-crafted features in deep learning models, challenging conventional wisdom in an attempt to refute the utility of manually-engineered features.
Of particular interest to this paper is the work by \newcite{ma2016end} where they introduce a strong end-to-end model combining a
bi-directional Long Short-Term Memory (Bi-LSTM) network with Convolutional Neural Network (CNN) character encoding in a Conditional Random Field (CRF).
Their model is highly capable of capturing not only word- but also character-level features. 
We extend this model by integrating an auto-encoder loss, allowing the model to take hand-crafted features as input and re-construct them as output, and show that, even with such a highly competitive model, incorporating linguistic features is still beneficial.
Perhaps the closest to this study is the works by \newcite{Ammar+:2014} and \newcite{Zhang+:2017}, who show how CRFs can be framed as auto-encoders in unsupervised or semi-supervised settings. 

With our proposed model, we achieve strong performance on the CoNLL 2003 English NER shared task with an $F_1$ of $91.89$, significantly outperforming an array of competitive baselines. We conduct an ablation study to better understand the impacts of each manually-crafted feature. Finally, we further provide an in-depth analysis of model performance when trained with varying amount of data and show that the proposed model is highly competent with only 60\% of the training set.

\section{Methodology}

In this section, we first outline the model architecture, then the manually crafted features, and finally how they are  incorporated into the model.

\subsection{Model Architecture}
\begin{figure}[tb]
\centering
\includegraphics[width=\columnwidth]{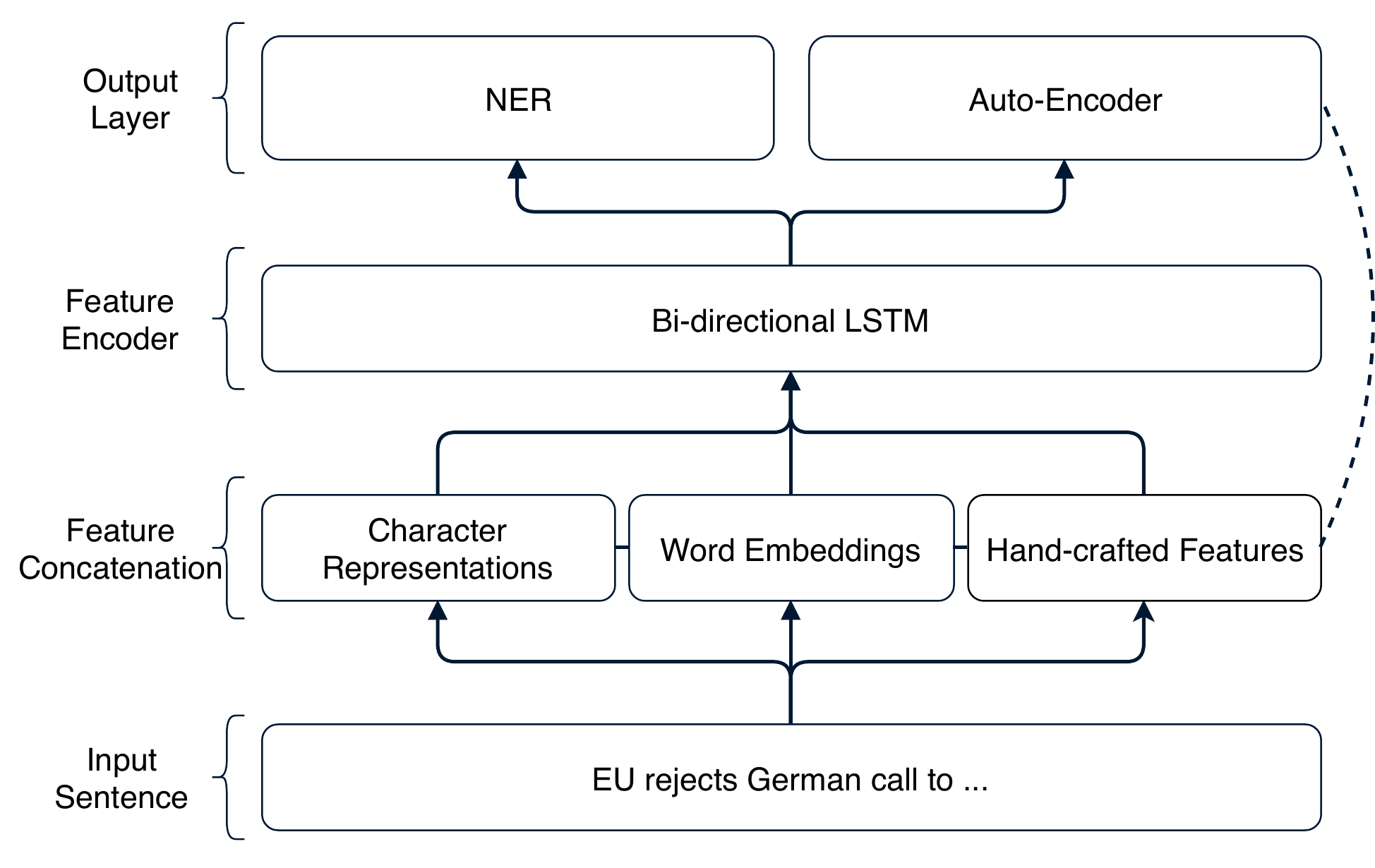}
\caption{Main architecture of our neural network. Character representations are extracted by a character-level CNN. The dash line indicates we use an auto-encoder loss to reconstruct hand-crafted features.}
\label{figure1}
\end{figure}

We build on a highly competitive sequence labelling model, namely Bi-LSTM-CNN-CRF, first introduced by \newcite{ma2016end}. 
Given an input sequence of $\mathbf{x} = \{x_{1}, x_2, \ldots, x_T\}$ of length $T$, the model is capable of tagging each input with a predicted label $\hat{y}$, resulting in a sequence of $\hat{\mathbf{y}} = \{\hat{y}_1, \hat{y}_2, \ldots, \hat{y}_T\}$ closely matching the gold label sequence $\mathbf{y} = \{y_1, y_2, \ldots, y_T\}$. 
Here, we extend the model by incorporating an auto-encoder loss taking hand-crafted features as in/output, thereby forcing the model to preserve crucial information stored in such features and allowing us to evaluate the impacts of each feature on model performance. 
Specifically, our model, referred to as Neural-CRF+AE, consists of four major components: (1) a character-level CNN (char-CNN); (2) a word-level bi-directional LSTM (Bi-LSTM); (3) a conditional random field (CRF); and (4) an auto-encoder auxiliary loss. 
An illustration of the model architecture is presented in \figref{figure1}.

\begin{table*}[!t]
\centering
\small
\begin{tabular}{llllllll}
\toprule
$\mathbf{x}$           & \em U.N.     & \em official & \em  Ekeus    & \em heads & \em for  & \em Baghdad & \em .     \\ 
\midrule
POS        & NNP      & NN       & NNP      & VBZ   & IN   & NNP     & .     \\ 
Word shape      & X.X.     & xxxx     & Xxxxx    & xxxx  & xxx  & Xxxxx   & .     \\ 
Dependency tags & compound & compound & compound & ROOT  & prep & pobj    & punct \\ 
Gazetteer       & O        & O        & PER      & O     & O    & LOC     & O     \\
\midrule
$\mathbf{y}$          & B-ORG      & O        & B-PER      & O     & O    & B-LOC     & O     \\ 
\bottomrule
\end{tabular}
\caption{Example sentence (top), showing the different types of linguistic features used in this work as additional inputs and auxiliary outputs (middle), and its labelling (bottom).}
\label{feature}
\end{table*}

\paragraph{Char-CNN.} 
Previous studies \cite{santos2014learning, chiu2016named, ma2016end} have demonstrated that CNNs are highly capable of capturing character-level features. 
Here, our character-level CNN is similar to that used in \newcite{ma2016end} but differs in that we use a ReLU activation \cite{nair2010rectified}.\footnote{While the hyperbolic tangent activation function results in comparable performance, the choice of ReLU is mainly due to faster convergence.} 

\paragraph{Bi-LSTM.} 
We use a Bi-LSTM to learn contextual information of a sequence of words. 
As inputs to the Bi-LSTM, we first concatenate the pre-trained embedding of each word $\vec[i]{w}$ with its character-level representation $\vec[w_i]{c}$ (the output of the char-CNN) and a vector of manually crafted features $\vec[i]{f}$ (described in \secref{sec:features}): 
\begin{align}
\vecfwd[i]{h} &= \lstmfwd(\vecfwd[i-1]{h}, [\vec[i]{w};\vec[w_i]{c};\vec[i]{f}])\\
\vecbwd[i]{h} &= \lstmbwd(\vecbwd[i+1]{h}, [\vec[i]{w};\vec[w_i]{c};\vec[i]{f}])\,,
\end{align}
where $[;]$ denotes concatenation.
The outputs of the forward and backward pass of the Bi-LSTM is then concatenated $\vec[i]{h} = [\vecfwd[i]{h};\vecbwd[i]{h}]$ to form the output of the Bi-LSTM, where dropout is also applied.

\paragraph{CRF.} 
For sequence labelling tasks, it is intuitive and beneficial to utilise information carried between neighbouring labels to predict the best sequence of labels for a given sentence. 
Therefore, we employ a conditional random field layer \cite{lafferty2001conditional} taking as input the output of the Bi-LSTM $\vec[i]{h}$. 
Training is carried out by maximising the log probability of the gold sequence: $\mathcal{L}_{CRF} = \log p(\mathbf{y}|\mathbf{x})$ while decoding can be efficiently performed with the Viterbi algorithm.

\paragraph{Auto-encoder loss.} 
Alongside sequence labelling as the primary task, we also deploy, as auxiliary tasks, three auto-encoders for reconstructing the hand-engineered feature vectors. 
To this end, we add multiple independent fully-connected dense layers, all taking as input the Bi-LSTM output $\vec[i]{h}$ with each responsible for reconstructing a particular type of feature: $\hat{\vec[i]{f}^{t}} = \sigma(\mat{W}^{t}\vec[i]{h})$ where $\sigma$ is the sigmoid activation function, $t$ denotes the type of feature, and $\mat{W}^{t}$ is a trainable parameter matrix. 
More formally, we define the auto-encoder loss as:
\begin{equation}
\mathcal{L}_{AE}^{t} = \sum_{i=0}^{T}\xentropy(\vec[i]{f}^{t}, \hat{\vec[i]{f}^{t}})\,.
\end{equation}

\myparagraph{Model training.}
Training is carried out by optimising the joint loss:
\begin{equation}
\mathcal{L} = \mathcal{L}_{CRF} + \sum_{t}\lambda^{t}\mathcal{L}_{AE}^{t}\,,
\end{equation}
where, in addition to $\mathcal{L}_{CRF}$, we also add the auto-encoder loss, weighted by $\lambda^t$. In all our experiments, we set $\lambda^t$ to $1$ for all $t$s.

\subsection{Hand-crafted Features}
\label{sec:features}

We consider three categories of widely used features: (1) POS tags; (2) word shape; and (3) gazetteers and present an example in \tabref{feature}.
While POS tags carry syntactic information regarding sentence structure, the word shape feature focuses on a more fine-grained level, encoding character-level knowledge to complement the loss of information caused by embedding lookup, such as capitalisation. 
Both features are based on the implementation of spaCy.\footnote{\url{https://spacy.io/}}
For the gazetteer feature, we focus on \textit{PERSON} and \textit{LOCATION} and compile a list for each. 
The \textit{PERSON} gazetteer is collected from U.S.\ census 2000, U.S.\ census 2010 and DBpedia whereas GeoNames is the main source for \textit{LOCATION}, taking in both official and alternative names. 
All the tokens on both lists are then filtered to exclude frequently occurring common words.\footnote{Gazetteer data is included in the code release.}
Each category is converted into a one-hot sparse feature vector $\vec[i]{f}^{t}$ and then concatenated to form a multi-hot vector $\vec[i]{f} = [\vec[i]{f}^{\textrm{POS}};\vec[i]{f}^{\textrm{shape}};\vec[i]{f}^{\textrm{gazetteer}}]$ for the $i$-th word. 
In addition, we also experimented with including the label of the incoming dependency edge to each word as a feature, but observed performance deterioration on the development set.
While we still study and analyse the impacts of this feature in \tabref{table4} and \secref{ssec:ablation}, it is excluded from our model configuration (not considered as part of $\vec[i]{f}$ unless indicated otherwise).

\section{Experiments}

In this section, we present our experimental setup and results for name entity recognition over the CoNLL 2003 English NER shared task dataset \cite{tjong2003introduction}. 

\subsection{Experimental Setup}
\paragraph{Dataset.} 
We use the CoNLL 2003 NER shared task dataset, consisting of 14,041/3,250/3,453 sentences in the training/development/test set respectively, all extracted from Reuters news articles during the period from 1996 to 1997. 
The dataset is annotated with four categories of name entities: \textit{PERSON, LOCATION, ORGANIZATION} and \textit{MISC}. 
We use the \texttt{IOBES} tagging scheme, as previous study have shown that this scheme provides a modest improvement to the model performance \cite{ratinov2009design, chiu2016named, lample2016neural, ma2016end}.

\paragraph{Model configuration.}
Following the work of \newcite{ma2016end}, we initialise word embeddings with GloVe \cite{pennington2014glove} ($300$-dimensional, trained on a 6B-token corpus).
Character embeddings are $30$-dimensional and randomly initialised with a uniform distribution in the range $[-\sqrt{\frac{3}{dim}}, +\sqrt{\frac{3}{dim}}]$.
Parameters are optimised with stochastic gradient descent (SGD) with an initial learning rate of $\eta = 0.015$ and momentum of $0.9$. 
Exponential learning rate decay is applied every 5 epochs with a factor of $0.8$. 
To reduce the impact of exploding gradients, we employ gradient clipping at $5.0$ \cite{pascanu2013difficulty}. 

We train our models on a single GeForce GTX TITAN X GPU.
With the above hyper-parameter setting, training takes approximately $8$ hours for a full run of $40$ epochs.

\paragraph{Evaluation.} 
We measure model performance with the official CoNLL evaluation script and report span-level named entity F-score on the test set using early stopping based on the performance on the validation set. 
We report average F-scores and standard deviation over 5 runs for our model.

\paragraph{Baseline.}
In addition to reporting a number of prior results of competitive baseline models, as listed in \tabref{table1}, we also re-implement the Bi-LSTM-CNN-CRF model by \newcite{ma2016end} (referred to as Neural-CRF in \tabref{table1}) and report its average performance.

\subsection{Results}
\begin{table}[tb]
\centering
\small
\begin{tabular}{lc}
\toprule
Model                     & $F_1$ \\
\midrule
\citet{chieu2002named} & 88.31 \\
\citet{florian2003named} & 88.76 \\
\citet{ando2005framework} & 89.31 \\
\citet{collobert2011natural}   & 89.59       \\
\citet{huang2015bidirectional} & 90.10       \\
\citet{passos2014lexicon}      & 90.90       \\
\citet{lample2016neural}       & 90.94       \\
\citet{luo2015joint}           & 91.20       \\
\citet{ma2016end}              & 91.21       \\
\citet{yang2017neural}         & 91.62 \\
\citet{Peters+:2018}           & 90.15 \\
\citet{Peters+:2018}+ELMo      & \textbf{92.22} ($\pm$ 0.10) \\
Neural-CRF$\ddagger$           & 91.06 ($\pm$ 0.18)  \\
Neural-CRF+AE$\ddagger*$       & 91.89 ($\pm$ 0.23)    \\ 
\midrule
\citet{ratinov2009design}$\dagger$ & 90.80       \\
\citet{chiu2016named}$\dagger$     & 91.62       \\ 
Neural-CRF+AE$\dagger$ $\ddagger$ & \textbf{92.29} ($\pm$ 0.20)\\
\bottomrule
\end{tabular}
\caption{NER Performance on the CoNLL 2003 English NER shared task test set. \textbf{Bold} highlights best performance. $\dagger$ marks models trained on both the training and development sets. $\ddagger$ indicates average performance over 5 runs. $*$ indicates statistical significance on the test set against Neural-CRF by two-sample Student's t-test at level $\alpha=0.05$.}
\label{table1}
\end{table}

The experimental results are presented in \tabref{table1}.
Observe that Neural-CRF+AE, trained either on the training set only or with the addition of the development set, achieves substantial improvements in F-score in both settings, superior to all but one of the benchmark models, highlighting the utility of hand-crafted features incorporated with the proposed auto-encoder loss.
Compared against the Neural-CRF, a very strong model in itself, our model significantly improves performance, showing the positive impact of our technique for exploiting manually-engineered  features.
Although \citet{Peters+:2018} report a higher F-score using their ELMo embedding technique, our approach here is orthogonal, and accordingly we would expect a performance increase if we were to incorporate their ELMo representations into our model.

\paragraph{Ablation Study}
\label{ssec:ablation}
\begin{table}[t]
\centering
\small
\begin{tabular}{lcc}
\toprule
Model                                            & Dev $F_1$ & Test $F_1$ \\
\midrule
Neural-CRF+AE & \textbf{94.87} ($\pm$ 0.21)   & \textbf{91.89} ($\pm$ 0.23)    \\
$-$ POS tagging$*$                                 & 94.78 ($\pm$ 0.17)   & 91.30 ($\pm$ 0.28)    \\
$-$ word shape$*$                                  & 94.83 ($\pm$ 0.31)   & 91.36 ($\pm$ 0.30)    \\
$-$ gazetteer                                    & 94.85 ($\pm$ 0.20)   & 91.80 ($\pm$ 0.19)    \\ 
$+$ dependencies                        & 94.74 ($\pm$ 0.16)   & 91.66 ($\pm$ 0.18)    \\
\bottomrule
\end{tabular}
\caption{Ablation study. Average performance over 5 runs with standard deviation. $+$ and $-$ denote adding and removing a particular feature (to/from Neural-CRF+AE trained on the training set only with POS tagging, word shape and gazetteer features). $*$ indicates statistical significance on the test set against Neural-CRF+AE by two-sample Student's t-test at level $\alpha=0.05$. Note that in this table, $*$ measures the drop in performance.}
\label{table4}
\end{table}

To gain a better understanding of the impacts of each feature, we perform an ablation study and present the results in \tabref{table4}.
We observe performance degradation when eliminating POS, word shape and gazetteer features, showing that each feature contributes to NER performance beyond what is learned through deep learning alone.
Interestingly, the contribution of gazetteers is much less than that of the other features, which is likely due to the noise introduced in the matching process, with many incorrectly identified false positives.

Including features based on dependency tags into our model decreases the performance slightly.
This might be a result of our simple implementation (as illustrated in Table~\ref{feature}), which does not include dependency direction, nor parent-child relationships.

\begin{table}
\centering
\small
\begin{tabular}{lcc}
\toprule
Model                                            & Dev $F_1$ & Test $F_1$ \\
\midrule
Neural-CRF & 94.53 ($\pm$ 0.21)   & 91.06 ($\pm$ 0.18)    \\
$+$ input                                  & 94.63 ($\pm$ 0.23)   & 91.17 ($\pm$ 0.25)    \\
$+$ output                               & 94.69 ($\pm$ 0.22)   & 91.23 ($\pm$ 0.19)    \\
$+$ input \& output$*$                                    & \textbf{94.87} ($\pm$ 0.21)   & \textbf{91.89} ($\pm$ 0.23)    \\ 
\bottomrule
\end{tabular}
\caption{Average performance of Neural-CRF with different features configurations over 5 runs with standard deviation. Note that $+$ input \& output = Neural-CRF+AE. $*$ indicates statistical significance on the test set against Neural-CRF by two-sample Student's t-test at level $\alpha=0.05$.}
\label{table5}
\end{table}

Next, we investigate the impact of different means of incorporating manually-engineered features into the model. 
To this end, we experiment with three configurations with features as: (1) input only; (2) output only (equivalent to multi-task learning); and (3) both input and output (Neural-CRF+AE) and present the results in \tabref{table5}.
Simply using features as either input or output only improves model performance slightly, but insignificantly so. 
It is only when features are incorporated with the proposed auto-encoder loss do we observe a significant performance boost.

\paragraph{Training Requirements}

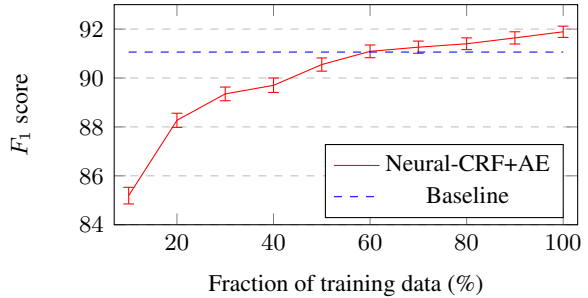
\begin{figure}[t]

\hspace{-1.5ex}
\begin{tikzpicture}
    \tikzstyle{every node}=[font=\small]

  \begin{axis}[
  	xlabel={Fraction of training data (\%)},
  	ylabel={$F_1$ score},
    xmin=7, xmax=103,
    ymin=84, ymax=93,
    height=4.5cm,
    ymajorgrids=true,
    grid style=dashed,
    width=0.48\textwidth,
    legend pos=south east
  ]

  \addplot[color=red, error bars/.cd, y dir=both, y explicit] coordinates {
  (10, 85.19)+=(10, 0.34)-=(10, 0.34)
  (20, 88.27)+=(20, 0.29)-=(20, 0.29)
  (30, 89.35)+=(30, 0.28)-=(30, 0.28)
  (40, 89.70)+=(40, 0.30)-=(40, 0.29)
  (50, 90.55)+=(50, 0.27)-=(50, 0.27)
  (60, 91.09)+=(60, 0.26)-=(60, 0.26)
  (70, 91.26)+=(70, 0.25)-=(70, 0.25)
  (80, 91.40)+=(80, 0.24)-=(80, 0.24)
  (90, 91.64)+=(90, 0.25)-=(90, 0.25)
  (100, 91.89)+=(100, 0.23)-=(100, 0.23)
  };
  \addlegendentry{Neural-CRF+AE}
  \addplot[blue, dashed] coordinates{(10,91.06)(100, 91.06)};
  \addlegendentry{Baseline}
  \end{axis}
\end{tikzpicture}

\vspace{-2ex}
\caption{Comparing the Neural-CRF+AE (red solid line) trained with varying amounts of data vs.\@ a Neural-CRF baseline (blue dashed line), trained on the full training set. Performance averaged over 5 runs, and error bars show $\pm$ 1 std.\@dev.}
\label{figure2}
\end{figure}

Neural systems typically require a large amount of annotated data.
Here we measure the impact of training with varying amount of annotated data,  as shown in  \figref{figure2}.
Wtih the proposed model architecture, the amount of labelled training data can be drastically reduced:
our model, achieves comparable performance against the baseline Neural-CRF, with as little as 60\% of the training data. 
Moreover, as we increase the amount of training text, the performance of Neural-CRF+AE continues to improve.

\paragraph{Hyperparameters}

\begin{figure}[t]

\hspace{-1ex}
\begin{tikzpicture}
    \tikzstyle{every node}=[font=\small]

  \begin{axis}[
  	xmode=log,
  	xlabel={$\lambda_i$},
  	ylabel={$F_1$ score},
    xmin=0, xmax=90,
    ymin=91.3, ymax=92,
    height=5.5cm,
    ymajorgrids=true,
    grid style=dashed,
    width=0.48\textwidth,
%    legend pos=south east
    legend style={at={(0.87,0)},anchor=south east}
  ]

  \addplot[color=red, error bars/.cd, y dir=both, y explicit] coordinates {
  (1e-8, 91.39)+=(1e-8, 0.0484)-=(1e-8, 0.0484)
  (1e-7, 91.43)+=(1e-7, 0.0361)-=(1e-7, 0.0361)
  (1e-6, 91.50)+=(1e-6, 0.04)-=(1e-6, 0.04)
  (1e-5, 91.56)+=(1e-5, 0.0529)-=(1e-5, 0.0529)
  (1e-4, 91.63)+=(1e-4, 0.0576)-=(1e-4, 0.0576)
  (1e-3, 91.62)+=(1e-3, 0.0484)-=(1e-3, 0.0484)
  (1e-2, 91.67)+=(1e-2, 0.0361)-=(1e-2, 0.0361)
  (1e-1, 91.74)+=(1e-1, 0.04)-=(1e-1, 0.04)
  (1, 91.89)+=(1, 0.0529)-=(1, 0.0529)
  (10, 91.56)+=(10, 0.0289)-=(10, 0.0289)
  };
  \addlegendentry{POS tagging}
  
  \addplot[color=blue, error bars/.cd, y dir=both, y explicit] coordinates {
  (1e-8, 91.41)+=(1e-8, 0.0289)-=(1e-8, 0.0289)
  (1e-7, 91.45)+=(1e-7, 0.0361)-=(1e-7, 0.0361)
  (1e-6, 91.52)+=(1e-6, 0.0324)-=(1e-6, 0.0324)
  (1e-5, 91.57)+=(1e-5, 0.04)-=(1e-5, 0.04)
  (1e-4, 91.66)+=(1e-4, 0.0441)-=(1e-4, 0.0441)
  (1e-3, 91.68)+=(1e-3, 0.0361)-=(1e-3, 0.0361)
  (1e-2, 91.70)+=(1e-2, 0.04)-=(1e-2, 0.04)
  (1e-1, 91.78)+=(1e-1, 0.0484)-=(1e-1, 0.0484)
  (1, 91.89)+=(1, 0.0529)-=(1, 0.0529)
  (10, 91.50)+=(10, 0.0324)-=(10, 0.0324)
  };
  \addlegendentry{Word Shape}
  
  \addplot[color=green, error bars/.cd, y dir=both, y explicit] coordinates {
  (1e-8, 91.72)+=(1e-8, 0.0225)-=(1e-8, 0.0225)
  (1e-7, 91.78)+=(1e-7, 0.0256)-=(1e-7, 0.0256)
  (1e-6, 91.76)+=(1e-6, 0.0324)-=(1e-6, 0.0324)
  (1e-5, 91.79)+=(1e-5, 0.0484)-=(1e-5, 0.0484)
  (1e-4, 91.80)+=(1e-4, 0.0576)-=(1e-4, 0.0576)
  (1e-3, 91.83)+=(1e-3, 0.0484)-=(1e-3, 0.0484)
  (1e-2, 91.85)+=(1e-2, 0.0361)-=(1e-2, 0.0361)
  (1e-1, 91.82)+=(1e-1, 0.04)-=(1e-1, 0.04)
  (1, 91.89)+=(1, 0.0529)-=(1, 0.0529)
  (10, 91.56)+=(10, 0.0289)-=(10, 0.0289)
  };
  \addlegendentry{Gazetteer}
  \end{axis}
\end{tikzpicture}
%}

\vspace{-2ex}
\caption{Effect of hyperparameter values on model performance. Each curve shows the effect of $\lambda_i$, for feature type $i$, with all other $\lambda_j=1, ~j \ne i$. Performance averaged over 5 runs, and error bars show $\pm$ 1 variance.}
\label{figure3}
\end{figure}
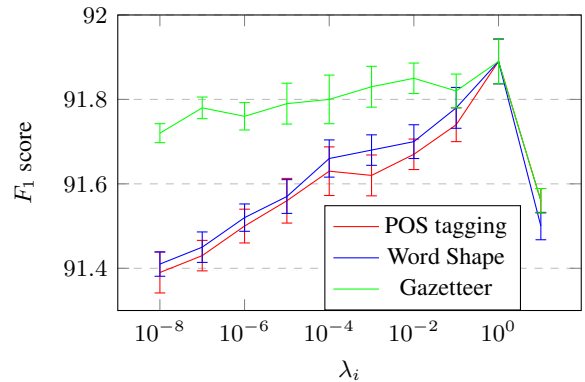

Three extra hyperparameters are introduced into our model, controlling the weight of the autoencoder loss relative to the CRF loss, for each feature type.
\figref{figure3} shows the effect of each hyperparameter on test performance. 
Observe that setting $\lambda_i=1$ gives strong performance, and that the impact of the gazetteer is less marked than the other two feature types. 
While increasing $\lambda$ is mostly beneficial, performance drops if the $\lambda s$ are overly large, that is, the auto-encoder loss overwhelms the main prediction task.

\section{Conclusion}
In this paper, we set out to investigate the utility of hand-crafted features. 
To this end, we have presented a hybrid neural architecture to validate this hypothesis extending a Bi-LSTM-CNN-CRF by incorporating an auto-encoder loss to take manual features as input and then reconstruct them. 
On the task of named entity recognition, we show significant improvements over a collection of competitive baselines, verifying the value of such features.
Lastly, the method presented in this work can also be easily applied to other tasks and models, where hand-engineered features provide key insights about the data.

\bibliographystyle{acl_natbib_nourl}
\bibliography{refer}

\end{document}